\documentclass{ecai} 
\usepackage{latexsym}
\usepackage{amssymb}
\usepackage{amsmath}
\usepackage{amsthm}
\usepackage{booktabs}
\usepackage{enumitem}
\usepackage{graphicx}
\usepackage{color}

\newcommand{\BibTeX}{B\kern-.05em{\sc i\kern-.025em b}\kern-.08em\TeX}

\begin{document}

%%%%%%%%%%%%%%%%%%%%%%%%%%%%%%%%%%%%%%%%%%%%%%%%%%%%%%%%%%%%%%%%%%%%%%%%

\begin{frontmatter}

%%% Use this command to specify your submission number.
%%% In doubleblind mode, it will be printed on the first page.

\paperid{1225} 

%%% Use this command to specify the title of your paper.

\title{Real-Time Indoor Object Detection based on hybrid CNN-Transformer Approach}

%%% Use this combinations of commands to specify all authors of your 
%%% paper. Use \fnms{} and \snm{} to indicate everyone's first names 
%%% and surname. This will help the publisher with indexing the 
%%% proceedings. Please use a reasonable approximation in case your 
%%% name does not neatly split into "first names" and "surname".
%%% Specifying your ORCID digital identifier is optional. 
%%% Use the \thanks{} command to indicate one or more corresponding 
%%% authors and their email address(es). If so desired, you can specify
%%% author contributions using the \footnote{} command.

\author[A,B]{\fnms{Salah-eddine}~\snm{LAIDOUDI}\thanks{Corresponding Author. Email: salah-eddine1.laidoudi@esme.fr}}
\author[B]{\fnms{Madjid}~\snm{MAIDI}}
\author[A]{\fnms{Samir}~\snm{OTMANE}} 

\address[A]{IBISC, Université Paris-Saclay, Univ Evry, Evry-Courcouronnes 91020, France }
\address[B]{ESME Research Lab, Ivry-sur-Seine, 94200, France}

%%% Use this environment to include an abstract of your paper.

\begin{abstract}
Real-time object detection in indoor settings is a challenging area of computer vision, faced with unique obstacles such as variable lighting and complex backgrounds. This field holds significant potential to revolutionize applications like augmented and mixed realities by enabling more seamless interactions between digital content and the physical world. However, the scarcity of research specifically fitted to the intricacies of indoor environments has highlighted a clear gap in the literature. To address this, our study delves into the evaluation of existing datasets and computational models, leading to the creation of a refined dataset. This new dataset is derived from OpenImages v7\cite{OpenImages}, focusing exclusively on 32 indoor categories selected for their relevance to real-world applications. Alongside this, we present an adaptation of a CNN detection model, incorporating an attention mechanism to enhance the model’s ability to discern and prioritize critical features within cluttered indoor scenes. Our findings demonstrate that this approach is not just competitive to existing state-of-the-art models in accuracy and speed but also opens new avenues for research and application in the field of real-time indoor object detection.
\end{abstract}

\end{frontmatter}

%%%%%%%%%%%%%%%%%%%%%%%%%%%%%%%%%%%%%%%%%%%%%%%%%%%%%%%%%%%%%%%%%%%%%%%%

\section{Introduction}
Object detection, a cornerstone of computer vision, has experienced transformative growth with the advent of deep learning technologies. Traditional techniques such as the Viola-Jones detector\cite{990517} and Histogram of Oriented Gradients (HOG)\cite{1467360} laid the early groundwork by enabling systems to recognize objects through feature detection and machine learning classifiers.

However, these methods struggled with high variability in object appearances and were generally limited in their ability to scale with complexity and diversity of input data.
The deep learning era introduced a paradigm shift with the development of convolutional neural networks (CNNs), significantly enhancing the ability and efficiency of object detection systems. This evolution continued with the introduction of advanced architectures, including single-stage detectors like YOLO (You Only Look Once)\cite{redmon2016look}\cite{redmon2016yolo9000}\cite{redmon2018yolov3}\cite{bochkovskiy2020yolov4}\cite{ultralytics2021yolov5}\cite{li2022yolov6}\cite{wang2022yolov7}\cite{Jocher_Ultralytics_YOLO_2023}\cite{wang2024yolov9}\cite{ge2021yolox} and SSD (Single Shot MultiBox Detector)\cite{Liu_2016}, and two-stage detectors such as R-CNN\cite{girshick2014rich} and its variants\cite{girshick2015fast}\cite{ren2016faster}, which improved detection accuracy by refining proposals through a secondary classification step. 

More recently, the integration of transformer-based models\cite{vaswani2023attention}, which employ self-attention mechanisms to capture global dependencies within the image data, has started to set new benchmarks in the field.

Despite these advancements, real-time object detection in indoor environments remains fraught with challenges. Indoor settings are characterized by varying lighting conditions—from natural light flooding a room to multiple artificial light sources—which can drastically affect the visibility and appearance of objects. Furthermore, object occlusions, where items are partially or fully hidden behind others, add an additional layer of complexity. Lastly, the demand for low latency in applications such as augmented reality (AR) and mixed reality (MR) means that the detection system not only needs to be accurate but also exceedingly fast. These conditions present unique hurdles that are not fully addressed by current object detection models.

The primary goal of our research is to develop an object detection system optimized for real-time application in indoor settings. This system is intended to significantly enhance user interactions in multiple fields such as augmented reality (AR) and mixed reality (MR), where seamless integration of digital content with the physical world is what makes these experiences more immersive.

Our work introduces a novel hybrid architecture that combines the robustness of CNNs with the sophisticated spatial reasoning capabilities of transformers. This approach is designed to be lightweight, catering to the needs of real-time processing without sacrificing accuracy. By employing this hybrid model, our system not only addresses the typical challenges associated with indoor object detection—such as variable lighting and occlusions—but also shows promising results that are competitive with, and in some cases superior to, current state-of-the-art technologies. Our contribution is a significant step forward in computer vision, pushing the boundaries of what is achievable in real-time indoor object detection.

%%%%%%%%%%%%%%%%%%%%%%%%%%%%%%%%%%%%%%%%%%%%%%%%%%%%%%%%%%%%%%%%%%%%%%%%

\section{Literature review}
\begin{figure*}[h!]
    \centering
    \includegraphics[width=0.8\linewidth]{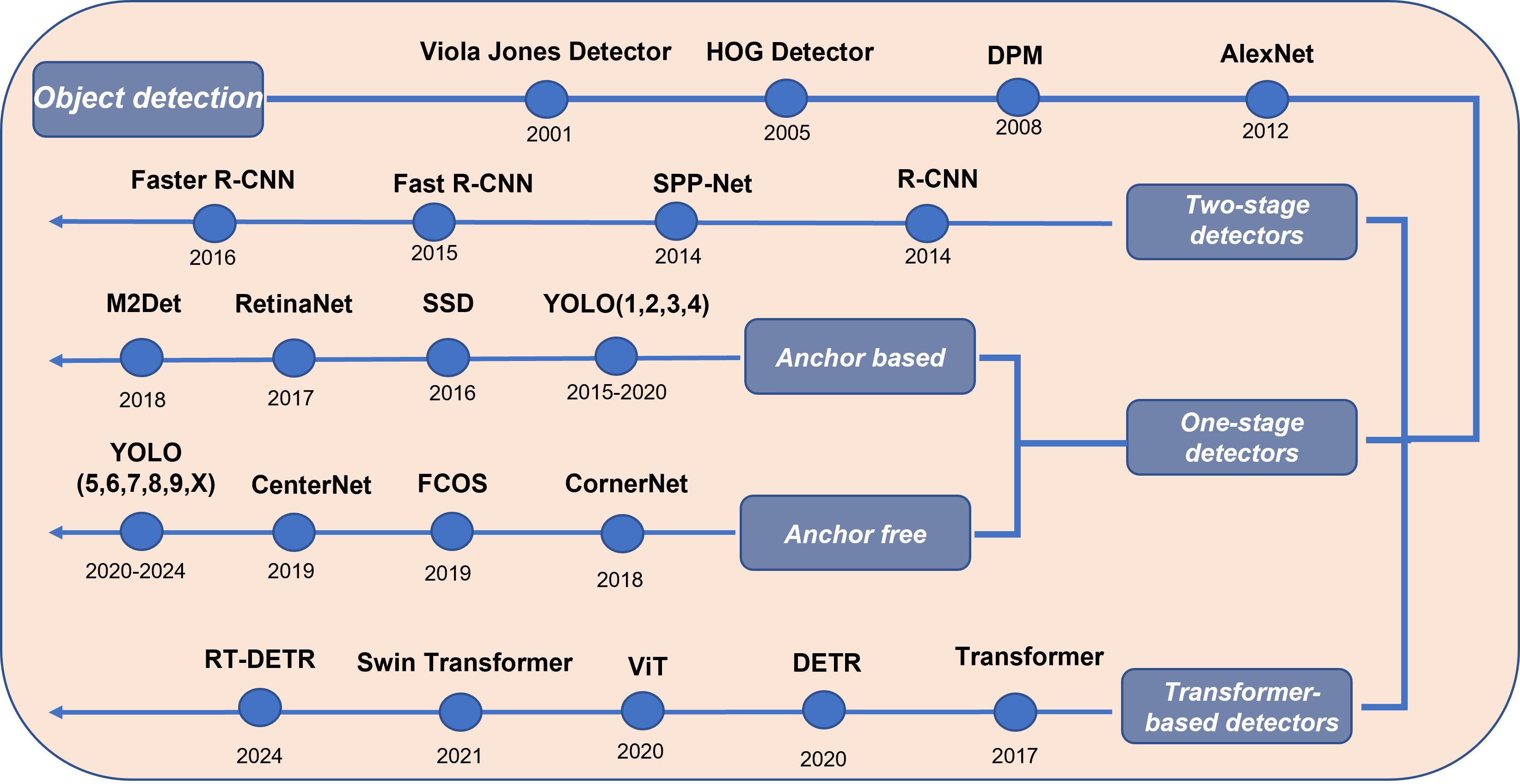}
    \caption{Most common object detection methods throughout the time}
    \label{fig:yolos}
\end{figure*}
Object detection has significantly evolved over the past decade due to advances in neural networks and machine learning. Despite this progress, research specifically targeting indoor environments is limited. Typically, object detectors are evaluated using the COCO dataset\cite{lin2015microsoft}, which is comprehensive but does not fully address the challenges of indoor settings, such as complex lighting, varied object placements, and diverse backgrounds. 

As we previously mentioned on the introduction, single-stage detectors have garnered attention for their efficiency, with a particular emphasis on speed, making them ideal candidates for real-time applications. Within this category, we observe two primary types of single-stage detectors:

\textbf{Anchor-Based Detectors:} These detectors, exemplified by earlier versions of YOLO, SSD, and RetinaNet ..., rely on predefined bounding boxes, known as anchors, which serve as priors to guide the detection process. They operate by pre-establishing a variety of bounding boxes with different widths and heights tailored to the common aspect ratios observed for each class type. During detection, these anchors are tiled across the image, and the model predicts the likelihood of an object being present within these preset tiles.

\textbf{Anchor-Free Detectors:} The recent shift towards anchor-free detectors signifies a promising trend in the field of real-time object detection, with notable examples including CornerNet\cite{law2019cornernet}, CenterNet\cite{duan2019centernet}, FCOS\cite{tian2019fcos}, and the latest iterations of YOLO (v5-v9). These modern detectors abstain from the reliance on prior bounding box knowledge. Instead, they directly predict critical points—specifically the top-left and bottom-right corners of the bounding box, thereby streamlining the detection process. This direct prediction approach circumvents the computational burden associated with anchor manipulation, leading to a more efficient detection framework suitable for rapid deployment in real-world scenarios.

Our research primarily aligns with the progressive trend of anchor-free detectors, we adopt and further this approach, augmenting it with the capabilities of a transformer.

Several noteworthy efforts in adjacent areas include innovations in low-textured object detection\cite{10337653}, where lighter SSD architectures have been proposed\cite{10398729}. For instance, a notable study suggested modifications to the SSD framework to better handle objects with minimal textural information, enhancing detection in specific but limited scenarios. Furthermore, the RT-DETR\cite{zhao2024detrs} model stands out as a pioneering real-time detection transformer, integrating the rapid inference capabilities necessary for real-time applications with the advanced spatial reasoning of transformers.

Additionally, the YOLO series  has significantly impacted the field with its high-speed processing and increasingly accurate detection rates. These models have consistently pushed the boundaries of what is achievable in real-time object detection, setting high benchmarks for both speed and accuracy. However, these systems are often calibrated and tested against datasets that do not adequately mimic the challenges specific to indoor environments.

In this last part of the literature review we’ll examine some of the models from our comparative study more closely. 

\textbf{Yolov5}\cite{ultralytics2021yolov5}, developed by Glen Jocher of Ultralytics in 2020 shortly after YOLOv4\cite{bochkovskiy2020yolov4}, is an advanced object detection model built on Pytorch rather than Darknet. It includes a pre-training tool called AutoAnchor that optimizes anchor boxes using a k-means function and a Genetic Evolution algorithm, which improves detection by evolving anchors over 1000 generations with CIoU loss and Best Possible Recall metrics. The architecture incorporates a modified CSPDarknet53 backbone, SPPF\cite{He_2014} (spatial pyramid pooling fast) for efficient feature processing at various scales, and various advanced augmentations like Mosaic, MixUp\cite{zhang2018mixup}, and HSV changes to enhance training. YOLOv5\cite{ultralytics2021yolov5} offers multiple model sizes (from nano to extra large) to accommodate different device capabilities and performance needs. It is open-source, supported by a large community, and includes tools for easy integration and deployment on mobile devices.

\textbf{YOLOv6} \cite{li2022yolov6}, developed by Meituan Vision AI Department and released in September 2022, advances real-time object detection with significant improvements in speed and accuracy over predecessors like YOLOv5\cite{ultralytics2021yolov5} and YOLOX\cite{ge2021yolox}. Its architecture features a new EfficientRep\cite{weng2023efficientrepan} backbone for enhanced parallel processing, a PAN \cite{liu2018path}topology neck, and an efficient decoupled head with hybrid-channel strategy. Notable innovations include Task Alignment Learning for label assignment, advanced VariFocal \cite{zhang2021varifocalnet} and SIoU/GIoU\cite{li2022yolov6} loss functions for precise detection, and optimized quantization techniques for faster performance. On the MS COCO dataset\cite{lin2015microsoft}, YOLOv6-L achieved an AP of 52.5\% and AP50 of 70\% at about 50 FPS on an NVIDIA Tesla T4, demonstrating its effectiveness in high-speed object detection applications.

 \textbf{YOLOv7}\cite{wang2022yolov7}, released in July 2022, significantly advances object detection with speeds ranging from 5 to 160 FPS. Developed using only the MS COCO dataset, it introduces key architectural improvements:

\begin{itemize}
    \item \textbf{Extended Efficient Layer Aggregation Network (E-ELAN):} Optimizes deep learning models by managing gradient paths efficiently.
    \item \textbf{Model Scaling for Concatenation-based Models:} Adjusts block depth and width to maintain optimal structure and efficiency.
\end{itemize}
YOLOv7 also features performance enhancements like revised re-parameterized convolution (RepConvN) for better network structure, dual label assignment for precision in training and output, and advanced batch normalization for improved inference efficiency. These innovations set new benchmarks in object detection technology.

\textbf{YOLOv8}\cite{Jocher_Ultralytics_YOLO_2023}, released by Ultralytics in January 2023, expands the YOLO family with versions ranging from nano to extra-large. It supports diverse vision tasks like detection, segmentation, and classification. Enhancements include an updated C2f module in the backbone for better feature integration and an anchor-free, decoupled head design that improves accuracy by processing objectness, classification, and regression separately. It employs advanced loss functions like CIoU\cite{9497076} and DFL\cite{hossain2020adaptive}, optimizing detection, especially for smaller objects.

Additionally, YOLOv8 introduces the YOLOv8-Seg for semantic segmentation, achieving top-tier results in benchmarks while maintaining high efficiency and speed. Available via CLI or as a PIP package, YOLOv8 offers easy integration for various applications. In testing, YOLOv8x reached an AP of 53.9\% and a speed of 280 FPS on an NVIDIA A100, surpassing previous models like YOLOv5.

\textbf{RT-DETR} or Real-Time DEtection TRansformer (RT-DETR)\cite{zhao2024detrs}, an innovative real-time end-to-end object detector that efficiently addresses the limitations of YOLO models and Transformer-based detectors like DETR\cite{carion2020endtoend}. RT-DETR optimizes performance by focusing first on enhancing speed without sacrificing accuracy, and then on improving accuracy while maintaining speed. Key advancements include an efficient hybrid encoder that rapidly processes multi-scale features, and an uncertainty-minimal query selection that enhances decoder accuracy.

RT-DETR also features adaptable speed tuning by varying the number of decoder layers, allowing customization to different operational scenarios without retraining. Performance tests on the COCO dataset show RT-DETR models achieving 53.1\% and 54.3\% AP for RT-DETR-R50 and RT-DETR-R101, respectively, surpassing prior models in both speed and accuracy. Moreover, post pre-training with Objects365\cite{9009553}, the models achieve up to 56.2\% AP. 

  Despite these developments, the field of indoor object detection remains underexplored. The predominant use of the COCO dataset for model evaluation introduces a significant limitation, as it does not adequately represent the specific challenges posed by indoor settings, such as varied lighting conditions and complex spatial arrangements. This mismatch between training environments and real-world applications indicates a critical gap in the literature.

Our research seeks to address these shortcomings by focusing specifically on real-time indoor object detection on a custom dataset made from the OpenImages v7 dataset. We aim to develop a model that not only meets the generic criteria of accuracy and speed but is also fine-tuned to the specific demands and conditions of indoor environments. By utilizing a novel hybrid architecture that combines the strengths of CNNs and transformers, our work is poised to offer new insights and substantial improvements over the current state-of-the-art methods, specifically made for real-world indoor applications.\\

%%%%%%%%%%%%%%%%%%%%%%%%%%%%%%%%%%%%%%%%%%%%%%%%%%%%%%%%%%%%%%%%%%%%%%%%

\section{Methodology}
\subsection{Data collection}

In this section, we'll describe the selection and preparation of a specialized subset of the OpenImages v7 dataset. We carefully curated 32 indoor object categories that are critical for understanding real-world indoor environments. This targeted dataset is designed to tackle specific challenges associated with indoor object detection, such as variable lighting and complex backgrounds.

To increase the robustness and diversity of our dataset, we incorporated the mosaic data augmentation technique. This technique constructs a single training image from four distinct images, enhancing the model's exposure to a variety of scenarios. It simulates complex interactions and occlusions between objects, which is vital for improving the model’s generalization ability across different indoor settings.

\begin{figure}[h!]
    \centering
    \includegraphics[width=0.8\linewidth]{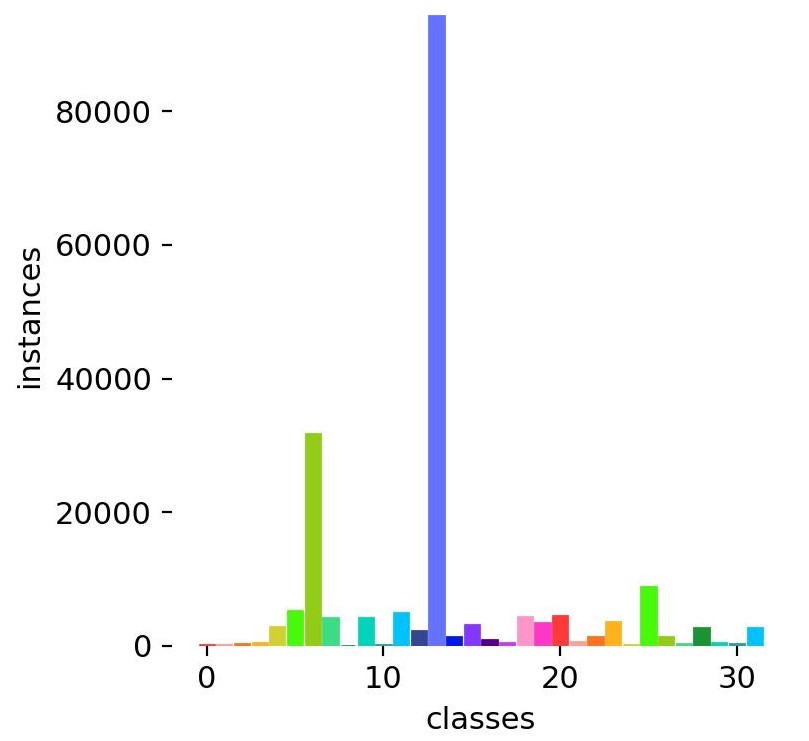}
    \caption{Classes distribution across the dataset}
    \label{fig:classes}
\end{figure}

Figure \ref{fig:classes} illustrates the distribution of class instances within the dataset. Notably, class 13 (chair) is more prevalent, resulting in a skewed distribution. However, we chose not to balance this class distribution in order to evaluate the effectiveness of the mosaic data augmentation\cite{chen2021dynamic} and DFL loss\cite{hossain2020adaptive} . These methods are intended to mitigate the impact of such imbalances on the model’s performance during final evaluations.

\begin{figure}
    \centering
    \includegraphics[width=0.75\linewidth]{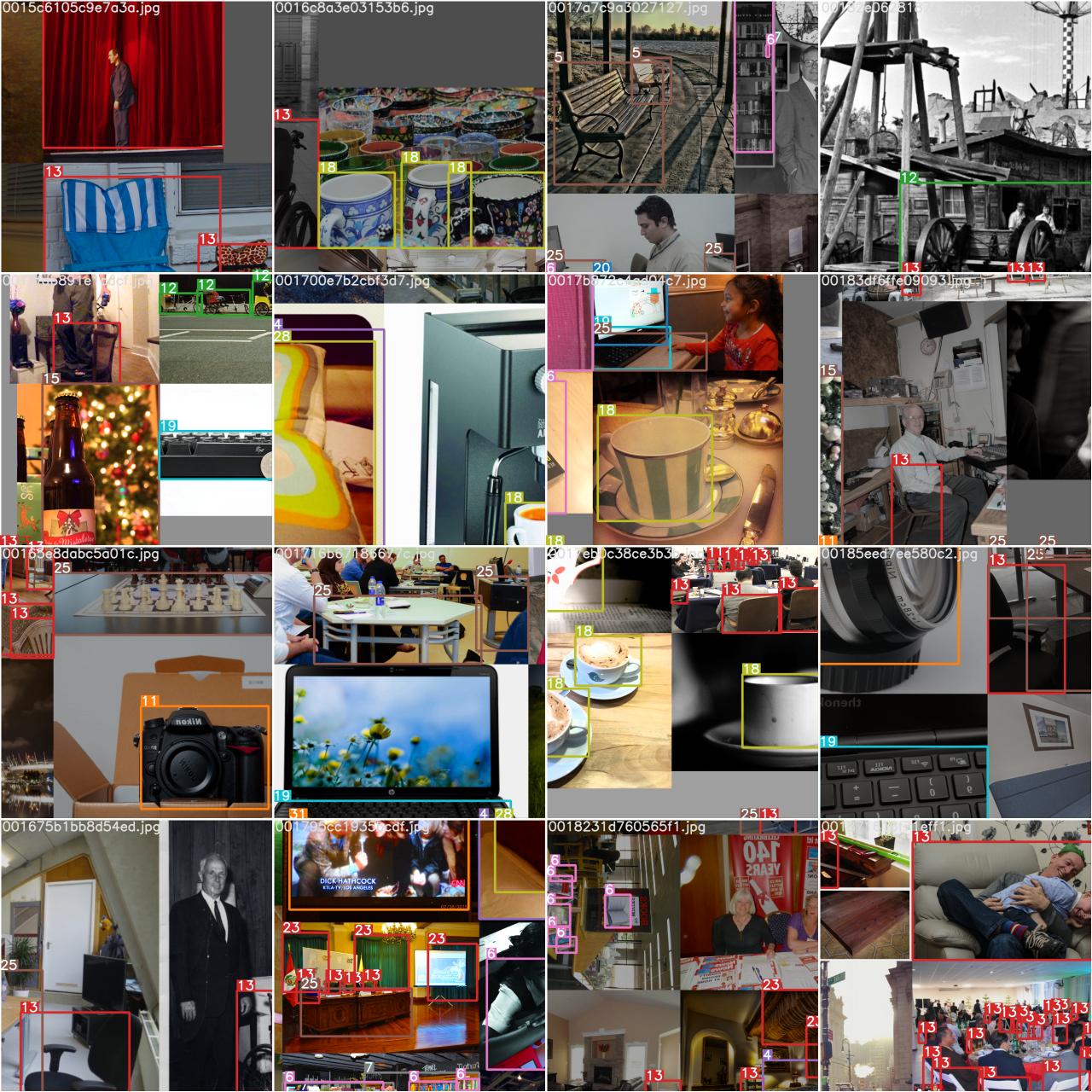}
    \caption{Training batch}
    \label{fig:batch}
\end{figure}
In Figure \ref{fig:batch} a training batch example is presented and the effect of the mosaic data augmentation technique can be clearly seen.
\subsection{Model Architecture}

In this section, we detail the architecture of our object detection model, which is a hybrid system combining Convolutional Neural Networks (CNNs) and Transformer-based models. This design is strategically chosen to effectively handle the spatial hierarchies and contextual dependencies typical in indoor environments.
\begin{figure*}[ht!]
    \centering
    \includegraphics[width=1\linewidth]{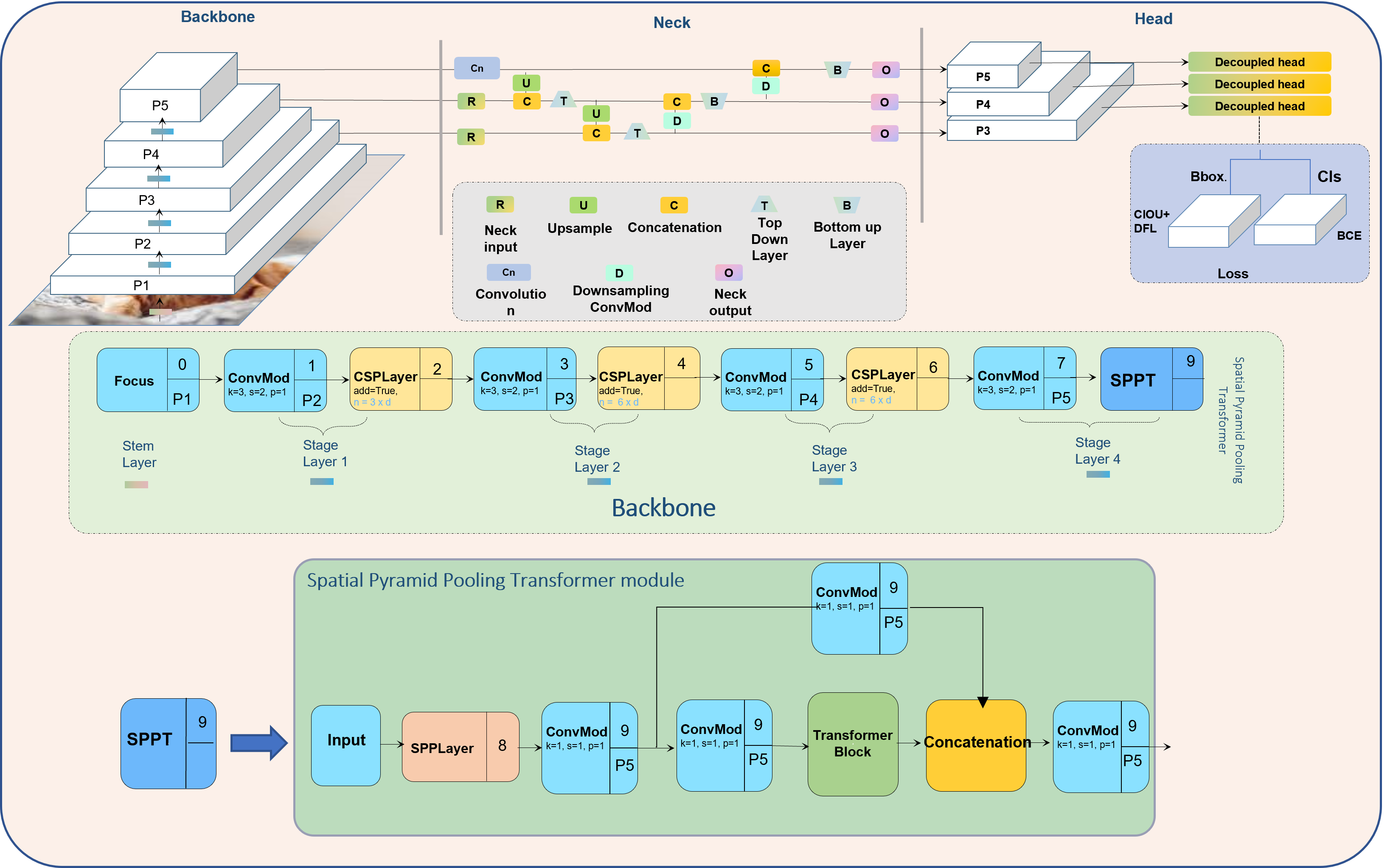}
    \caption{The general architecture of our model}
    \label{fig:architecture}
\end{figure*}
The CNN component of our model serves as a high-performance feature extractor, essential for recognizing and delineating object features at various scales. In contrast, the Transformer component of the architecture takes these extracted features and integrates them across the entire image. This integration allows the model to dynamically focus on areas of interest, leveraging the local processing strengths of CNNs with the global perspective capabilities of Transformers. This synergy aims to provide a robust solution for the complexities of indoor object detection.

Our model’s architecture is inspired by the YOLOv8n framework but is uniquely constructed from scratch using PyTorch. This approach gives us the flexibility to innovate without the constraints of the Ultralytics library.

A key modification in our model is the replacement of the conventional convolution layer at the input with a Focus layer. This layer concentrates spatial information from the input tensor into a channel-rich tensor, effectively doubling the channel capacity while reducing spatial dimensions. This configuration aids the network in learning spatial hierarchies more efficiently.
 The ConvMod ( Convolutional Module ) is a conv2d layer followed by a batch normalization layer and a SiLU activation function.

Further enhancements include the integration of a custom module, the Spatial Pyramid Pooling Transformer (SPPT). This module combines a conventional spatial pyramid pooling (SPP) layer with a Transformer block inspired by the Vision Transformer (ViT)\cite{dosovitskiy2021image}.  The introduction of the SPPT block has significantly reduced the model's floating-point operations (FLOPs), enhancing its computational efficiency and speed.  

The SPPT module follows the backbone’s final block and employs a standard Transformer layer to capture global information. The Transformer Encoder within this block features a Multi-Head Self-Attention Mechanism that updates and combines features from different subspaces for linear projection. This mechanism helps in capturing contextual information across the entire image and minimizing the loss of global features. The output through a multi-layer perceptron (MLP) ensures the non-linear enhancement of the self-attention mechanism's expressiveness.

Additionally, we draw inspiration from the YOLOv5 architecture’s C3 module, which uses convolutional layers and bottleneck modules to reduce training parameters and computation. Our custom CSPlayer mimics these features but innovates by incorporating the Focus layer within the bottleneck structures rather than at the initial convolution layer.

This architectural design is depicted in Figure \ref{fig:architecture}, illustrating both the individual components and their integration, highlighting how each contributes to efficient and accurate object detection within challenging indoor environments.

The loss structure consists of Binary Cross Entropy (BCE) for the classification branch and combines Distribution Focal Loss (DFL) and Complete Intersection over Union (CIoU) loss for the regression branch. 

DFL targets the class imbalance and improves accuracy in predicting bounding boxes, especially for objects with ambiguous boundaries by estimating the probability distribution of bounding box coordinates. CIoU loss enhances accuracy by considering the aspect ratios and overlap of the predicted and actual boxes.

The Binary Cross-Entropy loss for a binary classification task is defined as:
\begin{equation}
    \text{BCE} = -\frac{1}{N} \sum_{i=1}^N \left[y_i \cdot \log(p_i) + (1 - y_i) \cdot \log(1 - p_i)\right]
\end{equation}
where \( N \) is the number of observations, \( y_i \) is the actual label (0 or 1), and \( p_i \) is the predicted probability of the observation being in class 1.

The Distribution Focal Loss, a variant that focuses more on difficult examples, is given by:
\begin{equation}
    \text{DFL} = -\frac{1}{N} \sum_{i=1}^N \alpha_i (1 - p_i^\gamma) \log(p_i)
\end{equation}
where \( p_i \) is the predicted probability, \( \alpha_i \) is a weighting factor for class imbalance, \( \gamma \) is a focusing parameter, and \( N \) is the number of observations.

The Complete Intersection over Union loss, which includes terms for overlap, distance, and aspect ratio, is defined as:
\begin{equation}
    \text{CIoU} = 1 - \text{IoU} + \frac{\rho^2(\text{b}_{\text{pred}}, \text{b}_{\text{true}})}{c^2} + \alpha \cdot \text{v}
\end{equation}
where:
\begin{itemize}
    \item \( \text{IoU} \) is the Intersection over Union,
    \item \( \rho(\text{b}_{\text{pred}}, \text{b}_{\text{true}}) \) is the Euclidean distance between the center points of the predicted and actual bounding boxes,
    \item \( c \) is the diagonal length of the smallest enclosing box covering both bounding boxes,
    \item \( \text{v} \) is an aspect ratio consistency term,
    \item \( \alpha \) is a trade-off parameter.
\end{itemize}
\subsection{Training Process}

Each model was trained for 200 epochs, or until convergence, which was monitored through performance on a validation set. The training was conducted on an NVIDIA RTX 4090, utilizing a custom learning rate schedule to optimize convergence speed and model accuracy. The batch size was set at 16 to balance the trade-off between memory usage and the granularity of gradient updates. This training setup was chosen to ensure that the model learns effectively from the augmented dataset, adapting to the varied and complex scenarios presented during the training phase.

These hyperparameters are shown in Table \ref{tab:my_label}

\begin{table}
    \centering
    \begin{tabular}{|c|c|} \hline 
         OS& Windows 11\\ \hline 
         CPU& I7 13700K\\ \hline 
         RAM& 32Gb\\ \hline 
         GPU& RTX 4090 24Gb\\ \hline 
         Epochs& 200\\ \hline 
         Batch size& 16\\ \hline 
         Optimizer& SGD\\ \hline 
         Learning rate& Lambda LR\\ \hline
    \end{tabular}
    \caption{Training environment and different hyper-parameters}
    \label{tab:my_label}
\end{table}
\subsection{Evaluation Metrics}

The performance of our models was evaluated using several key metrics: accuracy, precision, recall, number of parameters and floating-point operations. Accuracy measures the proportion of correct predictions (both true positives and true negatives) among all evaluations. Precision and recall provide insights into the model’s capability to classify indoor objects correctly without overfitting to frequent labels or missing less common objects. Number of parameters and FlOPs are decisive for real-time applications, as they determine the feasibility of deploying these models in scenarios where rapid object detection is essential. These metrics collectively help in assessing the effectiveness and efficiency of our proposed model in real-world indoor settings.

Average Precision (AP) for a single class is calculated from the area under the precision-recall curve:
\begin{equation}
    \text{AP} = \int_0^1 p(r) \, dr
\end{equation}

Mean Average Precision (mAP) is the mean of the average precision scores for each class, widely used to evaluate object detection models:
\begin{equation}
    \text{mAP} = \frac{1}{N} \sum_{i=1}^N AP_i
\end{equation}
where \( N \) is the number of classes and \( AP_i \) is the average precision for class \( i \).

\section{Results and discussion}
\begin{figure}[h!]
    \centering
    \includegraphics[width=1\linewidth]{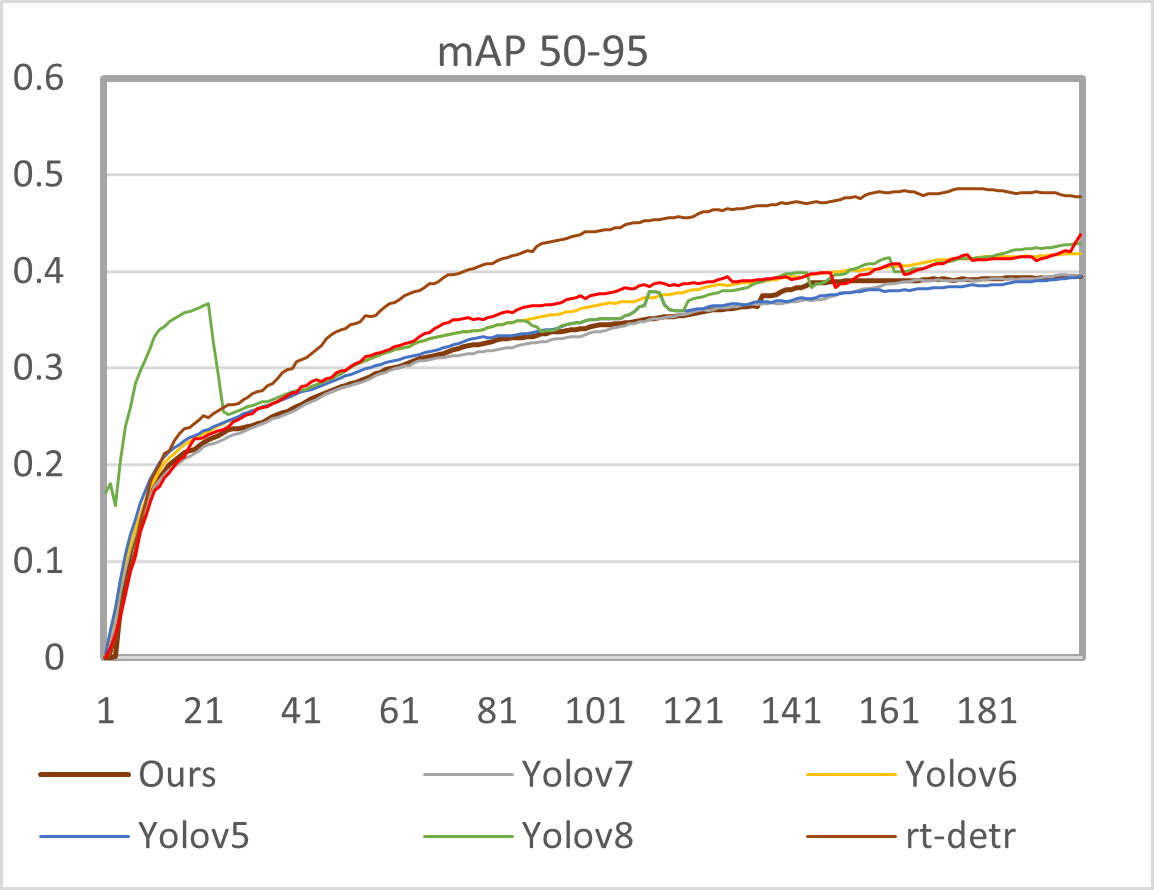}
    \caption{mAP50-95 evolution of all models across 200 epochs}
    \label{fig:mAP}
\end{figure}

In our discussion of the results presented in the graph shown in Figure \ref{fig:mAP} and Table \ref{Comparaison of different models}, we analyze the performance metrics across various models, focusing primarily on the mean Average Precision (mAP) and other key metrics such as recall and precision. 

\begin{table*}[h!]
    \centering
\caption{Comparaison of different models}
\label{Comparaison of different models}
    \scalebox{1}{
    \begin{tabular}{|c|c|c|c|c|c|c|c|c|c|} \hline 
         Model&  Year&  Size&  Param(M)&  FLOPs(B)&Precision&Recall & mAP50&  mAP50/95&Inference time(ms)\\ \hline 
         \textbf{Ours}&2024&640&2.772&4.4&0.63&0.509&0.54&0.388&12.2\\  
         Yolov8-rtdetr&  2024&  640& 9.545&16.9&0.62&0.506& 0.595&0.403&134\\
         RT-DETR&  2023&  640&  32.04&  103.6& 0.677&0.609& 0.628&0.467&184.5\\ 
         Yolov8n (pretrained on COCO)&  2023&  640&  3.2&  8.7& 0.625& 0.55& 0.582&  0.414&13.4\\ 
         Yolov8n (from scratch)&  2023&  640&  3.2&  8.7& 0.614& 0.538& 0.575&  0.407&19.9\\ 
         Yolov7 tiny&  2022&  640&  6.09&  13.3& 0.606& 0.562& 0.571&  0.376&15.9\\ 
         Yolov6n&  2022&  640&  4.237&  11.8& 0.614& 0.533&0.555&  0.397&15.2\\ 
         Yolov5n&  2020&  640&  2.509&  7.1& 0.618& 0.52& 0.554&  0.384&16.3\\ 
         \hline
    \end{tabular}
    }
    
\end{table*}
The graph shown in Figure\ref{fig:mAP} illustrates the evolution of the mAP over training epochs, where it becomes evident that while all models exhibit convergence, their performance in terms of precision across different classes is quite similar. This similarity in mAP values, ranging between 0.375 and 0.410 for most models, underscores a general consistency in model performance. Notably, the RT-DETR model slightly leads with a mAP of 0.47, a marginal but significant upper hand which can be attributed to its considerably larger model size compared to its counterparts.
Further reinforcing these findings, the results for recall and precision metrics also reveal minimal variability among the models, suggesting comparable effectiveness in identifying and accurately classifying objects within their respective datasets. This consistency is key for practical applications where predictability and reliability are valued.
\begin{figure}[h!]
    \centering
    \includegraphics[width=1\linewidth]{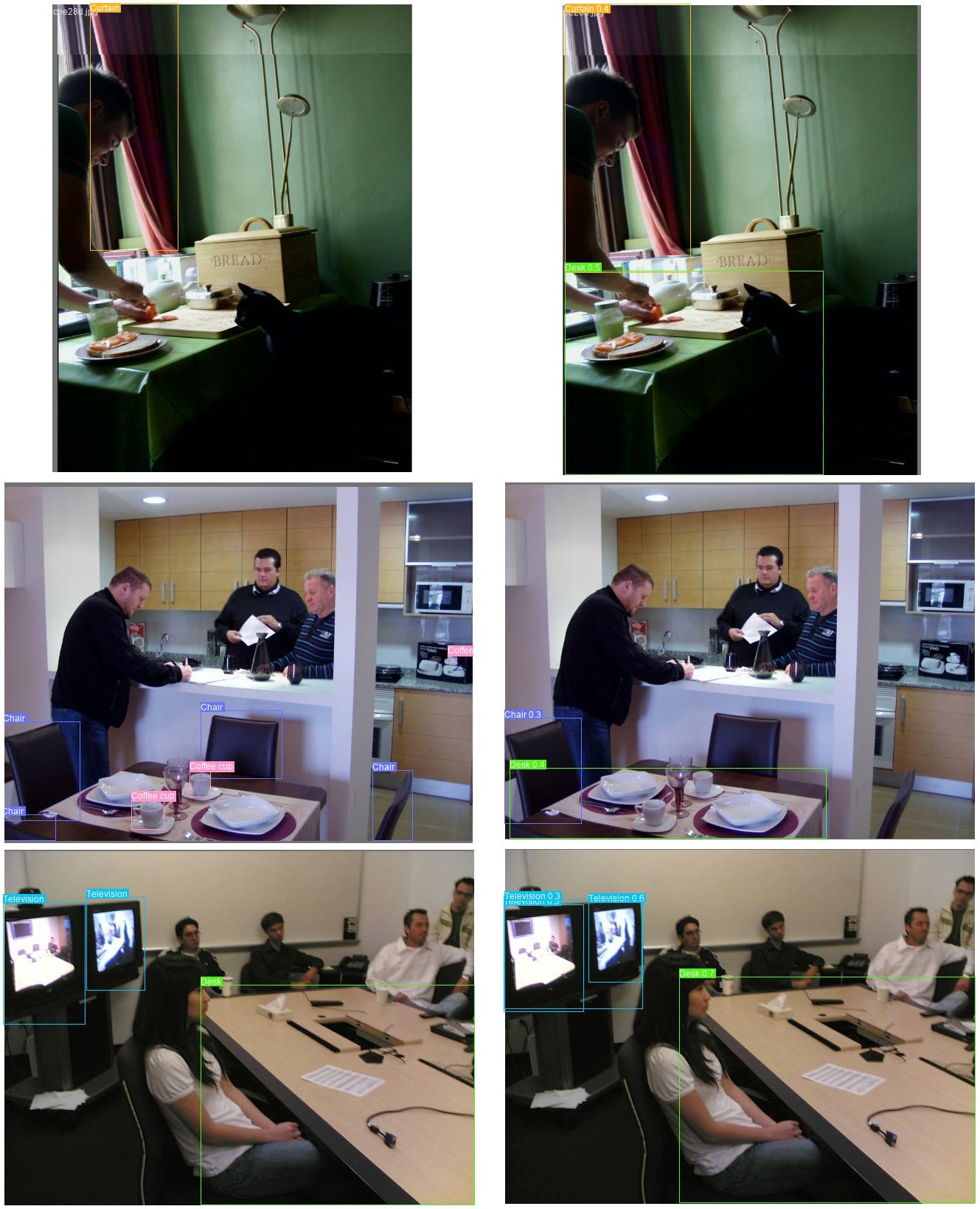}
    \caption{Detection comparison between ground truth (left) and our model's detection (right).}
    \label{fig: detection comparison}
\end{figure}

Moreover, Table \ref{Comparaison of different models} offers a detailed comparison of the models in terms of computational efficiency and speed. Our model stands out as the fastest, evidenced by its fewer parameters and lower computational complexity, measured in GFLOPs. This efficiency translates directly into enhanced performance, particularly when considering the inference time tested on an RTX 2070 Mobile GPU. The data clearly show that our model not only maintains competitive accuracy but also excels in operational speed, making it particularly suited for real-time applications where rapid processing is paramount.
\begin{figure}[h!]
    \centering
    \includegraphics[width=0.99\linewidth]{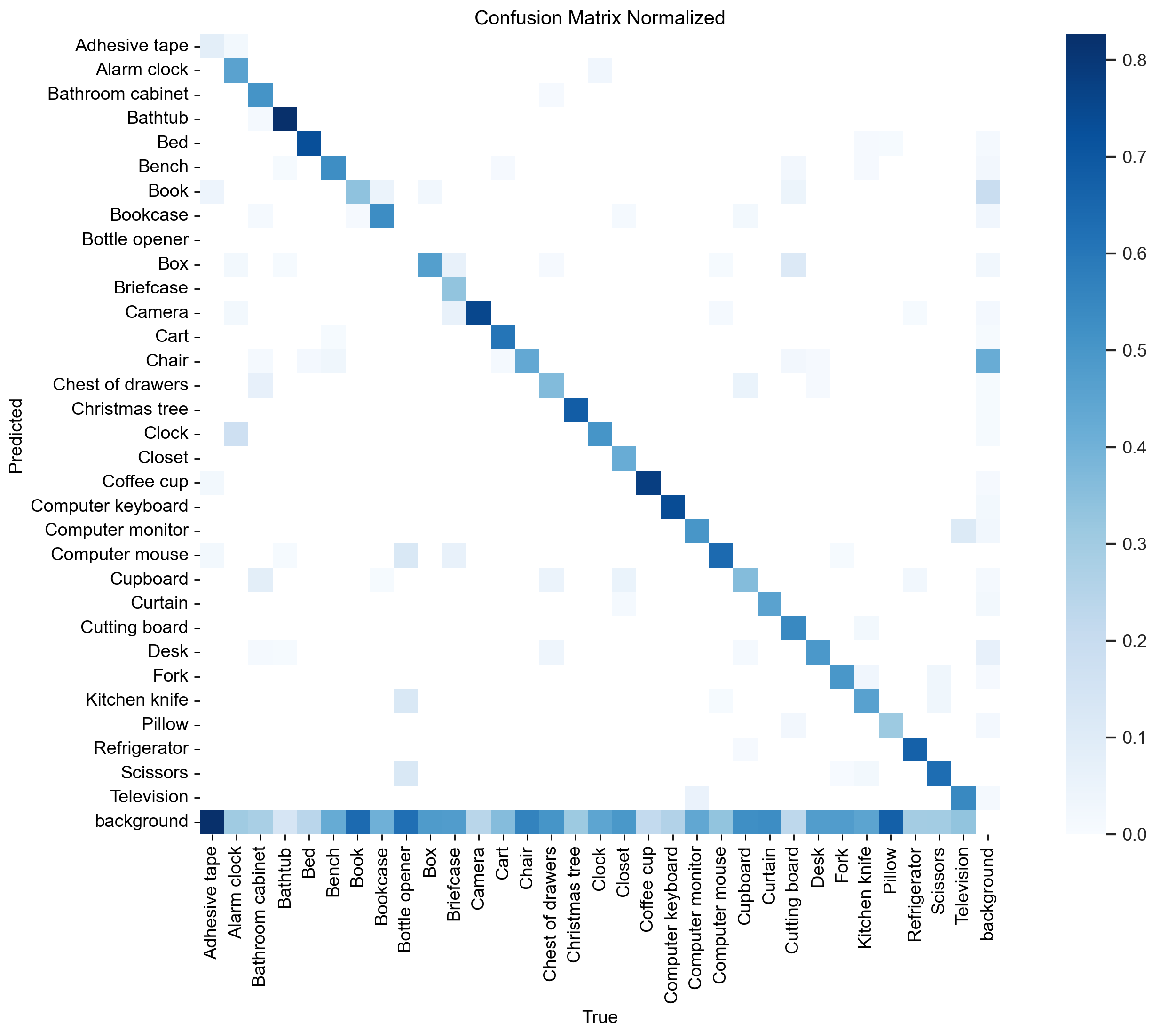}
    \caption{Confusion matrix of our model}
    \label{fig:matrix}
\end{figure}
These observations highlight the trade-offs between model complexity and speed, suggesting that while larger, more complex models like RT-DETR might achieve slightly higher accuracy, the increased computational demand may not always justify the marginal gains in performance. Our model's balance of speed and precision offers a compelling alternative, particularly for deployment in environments where both factors are critical. This balance will be a basis for further optimization and development aimed at refining our approach to indoor object detection, potentially extending its applicability to more demanding real-time environments.

In Figure \ref{fig:matrix}, we present the confusion matrix for our model, providing a visual and quantitative analysis of its performance across various classes within our dataset. The matrix reveals areas where the model excels, accurately predicting certain classes, while also identifying classes where performance is less satisfactory. A primary factor contributing to these disparities is the uneven representation of classes within the training data. Classes that are underrepresented tend to be more challenging for the model to learn, resulting in a lower predictive accuracy.

Despite these challenges, the overall performance of our model is deemed commendable. The strengths demonstrated in well-represented classes indicate the model's capability to learn and generalize from the data provided.

Figure \ref{fig: detection comparison} offers a qualitative comparison between our model's detection outcomes and the ground truth for bounding box annotations, across three randomly selected images from the test subset of our dataset. On the left side of the figure, we depict the ground truth bounding boxes, which serve as the benchmark for accurate object localization and classification. On the right, we display the corresponding detections made by our model, allowing for a direct visual assessment of its performance.

\section{Conclusion and future works}

In conclusion, our work in real-time indoor object detection marks significant advancements in the field, particularly through the creation and utilization of a customized dataset derived from the expansive OpenImages dataset. Our model, a hybrid CNN-Transformer  architecture, demonstrates the effectiveness of integrating these powerful technologies. We rigorously trained and compared various state-of-the-art real-time object detection models to our newly developed model. While the precision of these models was comparably high, our model stood out due to its lighter architecture, which substantially enhances processing speed. This feature is especially valuable in applications such as augmented reality (AR) and mixed reality (MR), where real-time processing on edge devices is pivotal. The efficiency and speed of our model not only meet these demanding requirements but also open avenues for further research and development across various real-time application scenarios, promising significant contributions to the practical deployment of different  technologies in everyday use 
\subsection{Future works :}

Despite the advancements achieved in our study, there remain several avenues for further enhancement and exploration in future work. Firstly, our model still relies on Non-Maximum Suppression (NMS) to refine detections, which can impact both speed and accuracy. To address this, we plan to explore end-to-end techniques like the one introduced by \cite{ouyang2024deyo} and their DEYO mode, such technique could eliminate the need for NMS, thereby simplifying the detection process and potentially improving performance metrics. Additionally, integrating advanced tracking algorithms such as SORT\cite{Bewley_2016} or DeepSORT \cite{wojke2017simple} will enhance the model's applicability in real-time scenarios, particularly in video surveillance and interactive systems where we need to maintain object consistency . Further development and expansion of our dataset will also be a priority, with a focus on including a wider variety of indoor scenarios to better train models to handle diverse lighting conditions, occlusions, and complex object interactions. Lastly, we aim to revisit and refine the loss function used during training to better accommodate the unique challenges posed by indoor object detection. This will involve tailoring the loss function to enhance precision in bounding box predictions and to effectively address issues related to class imbalance. These improvements will help to solidify the foundation for more robust and accurate object detection systems, particularly for applications on edge devices.

%%%%%%%%%%%%%%%%%%%%%%%%%%%%%%%%%%%%%%%%%%%%%%%%%%%%%%%%%%%%%%%%%%%%%%%%

%%% Use this environment to include acknowledgements (optional).
%%% This will be omitted in doubleblind mode.

%\begin{ack}
%By using the \texttt{ack} environment to insert your (optional) 
%acknowledgements, you can ensure that the text is suppressed whenever 
%you use the \texttt{doubleblind} option. In the final version, 
%acknowledgements may be included on the extra page intended for references.
%\end{ack}

%%%%%%%%%%%%%%%%%%%%%%%%%%%%%%%%%%%%%%%%%%%%%%%%%%%%%%%%%%%%%%%%%%%%%%%%

%%% Use this command to include your bibliography file.

\bibliography{m1225}

\end{document}